\documentclass[letterpaper, 10 pt, conference]{ieeeconf}  

\IEEEoverridecommandlockouts                              

\usepackage{graphics} 
\usepackage{epsfig} 
\usepackage{times} 
\usepackage{amsmath} 
\usepackage{amssymb}  
\usepackage{makecell}
\usepackage{multirow}
\usepackage{hyperref}

\usepackage{mathtools}
\usepackage{mathrsfs}
\usepackage{hyperref}
\usepackage{float}
\usepackage{pifont}
\usepackage{xcolor}
\usepackage{booktabs} 
\usepackage{graphicx}
\usepackage{threeparttable} 
\usepackage{tabularx}

\usepackage{fancyhdr}
\fancypagestyle{firstpage}{ 
    \fancyhf{} 
    \fancyhead[L]{Accepted to IEEE International Conference on Robotics and Automation (ICRA) 2025} 
}

\title{\LARGE \bf
Automatic Robotic-Assisted Diffuse Reflectance Spectroscopy Scanning System
}

\author{
        Kaizhong Deng$^{1,2}$,
        Christopher J. Peters$^{2}$,
        George P. Mylonas$^{1,2}$,
        Daniel S. Elson$^{1,2}$ %
\thanks{$^{1}$Hamlyn Centre for Robotic Surgery, Institute of Global Health Innovation, Imperial College London}%
\thanks{$^{2}$Department of Surgery and Cancer, Imperial College London, Exhibition Road, London, SW7 2AZ, UK. Corresponding email: {\tt\small daniel.elson@imperial.ac.uk}}%
}

\begin{document}

\maketitle

\thispagestyle{firstpage}

\pagestyle{empty}

\begin{abstract}

Diffuse Reflectance Spectroscopy (DRS) is a well-established optical technique for tissue composition assessment which has been clinically evaluated for tumour detection to ensure the complete removal of cancerous tissue. While point-wise assessment has many potential applications, incorporating automated large-area scanning would enable holistic tissue sampling with higher consistency. We propose a robotic system to facilitate autonomous DRS scanning with hybrid visual servoing control. A specially designed height compensation module enables precise contact condition control. The evaluation results show that the system can accurately execute the scanning command and acquire consistent DRS spectra with comparable results to the manual collection, which is the current gold standard protocol. Integrating the proposed system into surgery lays the groundwork for autonomous intra-operative DRS tissue assessment with high reliability and repeatability. This could reduce the need for manual scanning by the surgeon while ensuring complete tumor removal in clinical practice.
\end{abstract}

\section{INTRODUCTION}

The full resection of a tumour, while preserving healthy tissue, is one of the most critical objectives for many types of cancer surgery, which is correlated to minimal local recurrence and prolonged disease-free survival~\cite{negative_CRM}. To ensure optimal results, it is necessary to guarantee that no tumour is present at the edges of the removed specimen. Thus, there is a demand for intra-operative tissue assessment tools capable of characterising tissue in real time.
 
Several optical sensing technologies have been shown to facilitate real-time, non-invasive tissue classification~\cite{Nazarian2022JAMA,raman3,pcle_review}. Diffuse Reflectance Spectroscopy (DRS) is one of the most well-established techniques for tissue assessment which has recently been shown to be accurate at distinguishing between normal and cancer in upper~\cite{Nazarian2022JAMA} and lower~\cite{Nazarian2024IJS} gastrointestinal tissues. It has the advantages of low cost, high accuracy, ease of use and real-time signal analysis. 

DRS operates by illuminating the tissue with a band of light and measures the reflected light after its interaction with the tissue through scattering and absorption~\cite{DRS_principle}. The analysis of the spectral response allows identification of the optical properties of the tissue, hence determining its composition. DRS is usually configured as a hand-held probe, allowing surgeons to manually position it at a target location and requires steady tissue contact to obtain high quality data. 

\begin{figure}[tbp]
\includegraphics[width=\linewidth]{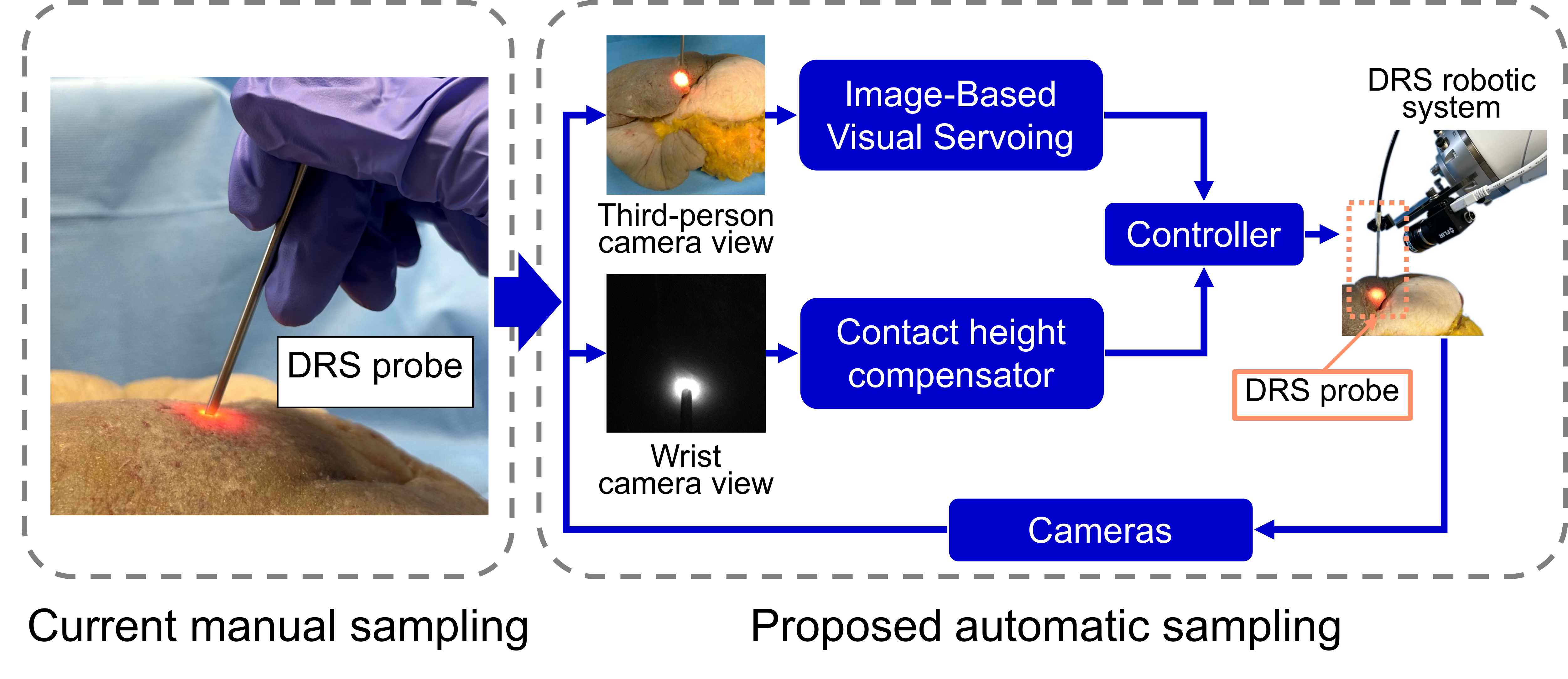}
\vspace{-20pt}
\caption{\textbf{Overview of the automatic DRS sampling system}: previous DRS sampling relies on manual operation to position the DRS probe. This proposed system utilised the visual servoing module and contact height compensator to the robot to automatically position the probe. Two cameras capture views from different scales as feedback to the system.}
\vspace{-10pt}
\label{fig:brief_overview}
\end{figure}
 
\begin{figure*}[!t]
\vspace{5pt}
\centering
\includegraphics[width=\linewidth]{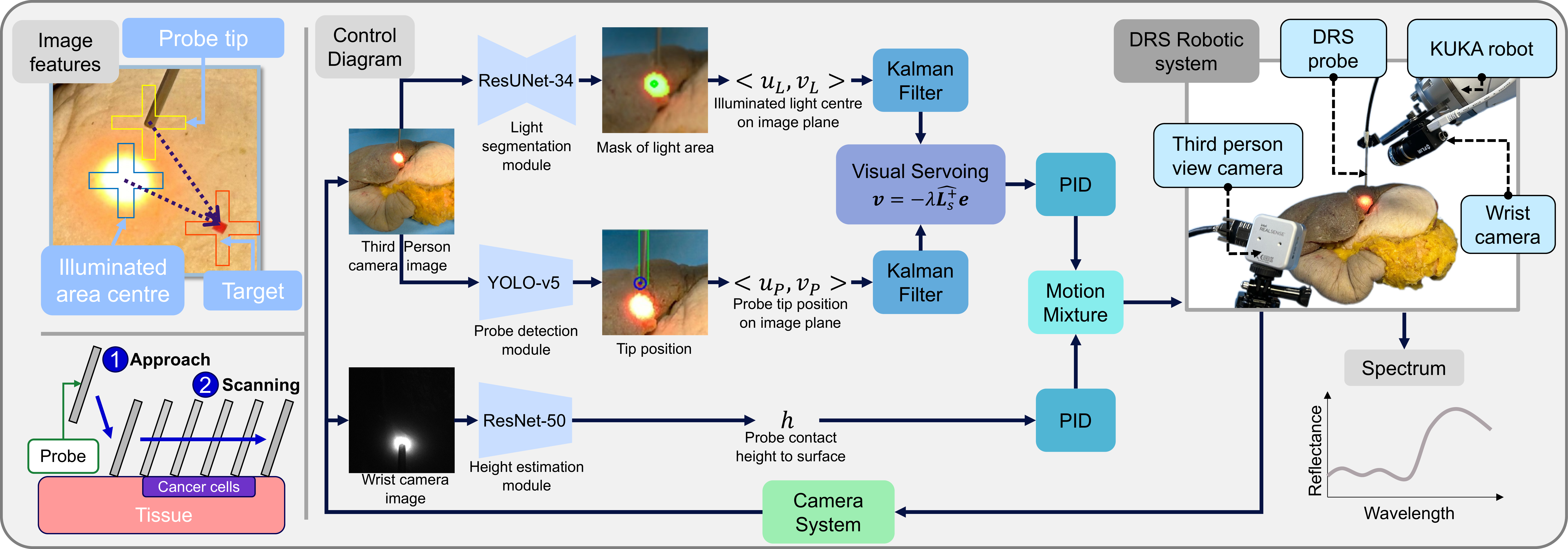}
\vspace{-20pt}
\caption{\textbf{DRS automatic system}: The left top is a zoomed view of a third-person camera image showing two image features: probe tip and illuminated area centre (also named as light centre), and a target point represented by a marker. The bottom left image illustrates the two stages of a DRS sampling procedure: the approach stage and the scanning stage. The right section is a diagram of the overall control. The system takes images from both cameras to extract the current visual state using two pairs of features and probe contact height. The visual servoing generates the motion to execute the scanning trajectory while tracking the target contact height. The resulting action drives the robot to sample DRS on the tissue surface.}
\label{fig:main_fig}
\end{figure*}
 
Despite the success for single-site sampling, the utility of large-area \textit{in vivo} sampling has not yet been realised.
A significant challenge of manual DRS collection is its high variability due to unsteady hand movement and variable contact pressure from the operator~\cite{DRS_pressure,DRS_pressure_2}.
For example, the contact pressure exhibits a non-linear relationship to the acquired spectrum~\cite{DRS_pressure_3}, which can cause different spectral responses for the same tissue type. This is also the main reason for the intra-observer variance between different operators, leading to unreliable results. An automated robot-assisted scanning system capable of controlling the contact condition could enhance the consistency of DRS in highly structured large-area tissue scanning scenarios. Such a system could standardise the DRS scanning procedure, generate highly standardised and trustworthy results, and free practitioners from tedious manual scanning. 

Previous research on robotic-assisted spectroscopy primarily focused on \textit{ex vivo} specimens.
For example, some robotic scanning systems require specimens to be prepared as microscope slides~\cite{scan_ex-vivo_plane_rob_ramen} or cut into smaller pieces~\cite{scan_rotation_rob_raman}.
However, these systems require meticulous specimen preparation, including dissection into slides and smaller samples, which is not feasible in a real-time surgical setting, making such methods impractical for use during surgery. Thus, \textit{in vivo} tissue scanning should accommodate to its original shape or size.
Recently, robotic systems have been developed for DRS scanning of large meat samples~\cite{DRS_scan_meat} and organ phantoms~\cite{DRS_scan_phantom}. However, these systems lack contact control to maintain constant contact, which might affect the results.

Visual servoing has been widely used to control robotic systems using images as feedback~\cite{VS_tutorial} and has been applied to medical robotic systems~\cite{VS_medical_robo_review}. For instance, it has enabled surgical robots to automatically position a laser instrument holder~\cite{VS_laser_instrument_dvss}, control the scanning of microscopic probes~\cite{scan_in_pCLE_kalman}, and control the endoscope~\cite{VS_lap_dvss}. Additionally, it has been used to control an instrument mounted on a robot arm for automatic middle ear surgery~\cite{VS_middle_ear} and for generating pCLE mosaics~\cite{VS_pcle_moasics_single_arm}. Thus, visual servoing is well-suited to position the DRS probe precisely with a robotic system. 

This study proposes a framework for a robotic-assisted autonomous DRS scanning system. It utilised a hybrid visual servoing approach with a precise height compensation module to acquire a consistent DRS spectrum as shown in \autoref{fig:brief_overview}. To the best of our knowledge, this is the first achievement of automatic robotic scanning using DRS with a contact condition management and without requiring any further tissue preparation steps. The primary contributions of this study are as follows:
\begin{itemize}
    \item Integration of the DRS instrument into a robotic system, controlled via visual servoing.
    \item Development of a precise height estimation and compensation module based on a deep neural network for real-time feedback.
    \item Evaluation of the system on two phantom tissue models and two animal tissue samples, demonstrating its trajectory precision and spectrum consistency to be comparable to manual collection, which is the current gold standard collection protocol.
\end{itemize}

\section{PROBLEM STATEMENT}
\textbf{Task description.} The sampling of DRS on a target surface consists of two stages: the approach stage and the in-contact scanning stage. During the approach stage, the DRS probe is moved to the start of a planned path on tissue. Once the firm contact has been established, the robot should switch to the in-contact scanning stage to automatically scan along a path indicated by the surgeon on the monitor of a third-person view camera image. The sampling should be made at the standard contact plane where the minimal pressure is made. The DRS spectrum collected along the scanned path is recorded during the in-contact scanning stage.

\textbf{Spectrum evaluation.} The reflectance spectrum of DRS can be denoted as $\mathbf{\mu}_{DRS}$. The quality of DRS spectrum is closely related to the performance of the system. The evaluation of $\mathbf{\mu}_{DRS}$ can be carried out from two perspectives, the intensity or the fingerprint of the spectra. 
The intensity ($\mathbf{\mu}_i$) represents the amount of light received during each sampling period.
The spectral fingerprint ($\mathbf{\mu}_f$) indicates the composition of the target, i.e., the signal received for each wavelength.
These two components can be represented as:
\begin{equation}
    \mathbf{\mu}_i = \sum_{\lambda}^{N_{\lambda}} \frac{\mathbf{\mu}_{DRS}(\lambda)}{N_{\lambda}},\ \mathbf{\mu}_f(\lambda) = \frac{\mathbf{\mu}_{DRS}(\lambda)}{\| \mathbf{\mu}_{DRS}(\lambda) \|_2} \, ,
\end{equation}
where $N_{\lambda}$ represent the total channel number of $\mathbf{\mu}_{DRS}(\lambda)$.

\textbf{Objectives.}  
The robotic scanning system is designed to achieve the following objectives while sampling on the same subject: (i) it should produce an intensity distribution comparable to manual collection, (ii) it should exhibit minimal differences in the spectral fingerprints, (iii) it should result in lower intra-observer variance than manual collection. 

\section{METHOD}

\subsection{Robotic Equipment Design}
We utilised a KUKA LBR iiwa14 R820 robot arm for this system. The DRS instrument was mounted on the robot with a custom-fit 3D-printed holder and consisted of a handheld reflection fiber housed within a stainless steel probe, a halogen light source, a visible-near infrared spectrometer, a monocular camera, and a data acquisition laptop. The detailed instrument setup has been described previously~\cite{Nazarian2022JAMA}. The monocular camera, for visual servoing, was set to look at the tip of the probe as a wrist camera. A RealSense D405 camera was fixed to the side of the robot acting as a third-person camera. The system was integrated using ROS and \texttt{iiwa\_stack} was employed to command the KUKA robot~\cite{iiwa_stack}. The overall system architecture is in \autoref{fig:main_fig}.

\subsection{Visual Servoing}
Image-based visual servoing (IBVS) was used to provide closed-loop control to guide the end-effector to the target point. IBVS aims to minimise the distance error of image features from their current position to the target position by iteratively approaching it~\cite{VS_tutorial}.
It uses two pairs of image features consisting of the position of the probe tip $\mathbf{s_p}=(u_p,v_p)^T$ and the centre of the illuminated light area $\mathbf{s_l}=(u_l,v_l)^T$ on the global camera images. The feature vector is thus defined as $\mathbf{s}(t)=(\mathbf{s_p}(t),\mathbf{s_l}(t))^T=(u_p(t),v_p(t),u_l(t),v_l(t))^T$. The error $\mathbf{e}$ reaches zero when the probe tip and the centre of the illuminated area reach the given target point. The velocity of the image features is represented by its derivative $\mathbf{\dot{s}}(t)$. The Cartesian velocity of the end-effector is annotated as $\mathbf{\dot{x}} = \mathbf{v}=(v_x,v_y,v_z)^T$. Only the position was controlled while the direction was set to always point downward.
The relationship between image feature space and Cartesian space is as $\mathbf{\dot{s}}=\mathbf{J}_{s} \mathbf{\dot{x}}$, where $\mathbf{J}_{s}$ is the Image Jacobian.  
To build a velocity controller to ensure an exponential decrease of the error $\mathbf{\dot{e}}=\mathbf{J}_{s} \mathbf{v} - \mathbf{\dot{s}}^*$, the velocity command could be then expressed as
$\mathbf{v} =-\lambda \mathbf{J}^+_s \mathbf{e}$,
where $\lambda$ is a positive gain and $\mathbf{J}^+_s$ is the Moore-Penrose pseudo-inverse. This closed-loop control enabled the end-effector to move smoothly towards the target.

\subsection{Tracking Probe Tip}\label{subsec:tip}
The position of the probe tip was detected using a YOLO-v5-based detection module as described in \cite{DRS_track}. It was trained on video data collected during \textit{ex vivo} DRS sampling in the operating theatre. To further smooth the detection results in this setting, a Kalman Filter was applied to reduce detection noise and estimate an even velocity.

\subsection{Tracking the Illuminated Area}\label{subsec:light}
To track the illuminated area, this region was first segmented and its centre was then tracked through the acquisition.
To collect segmentation training data, we manually directed the robot toward the contact surface and then used it to scan a liver phantom and a lamb liver.  
A total of 640 images were used for training while 160 images were used for validation. A customised UNet~\cite{unet} with a ResNet-34~\cite{resnet} as an encoder was used to generate the segmentation mask. 
After training for 200 epochs with a learning rate of $1\times10^{-4}$ using the Adam optimiser and BCE loss, the validation results showed a Dice loss of 0.0847 and a Jaccard loss of 0.155, which were sufficient for real world testing.

In the deployment, the centre of the illuminated area was extracted using morphological post-processing and centre-of-mass calculation.
A Kalman Filter was implemented to filter and interpolate the results, which could effectively reduce noise and speed up the prediction from $15\, \text{Hz}$ to $30\, \text{Hz}$. 

\subsection{Image Jacobian Estimation}
The estimated image Jacobian could more effectively represent the global relationship between visual features and the robot movement than analytical models.
Despite the advantage of online image Jacobian estimation~\cite{VS_M-Estimator,VS_kalman,VS_KNN-LLS}, an offline estimation pipeline was adopted for its safety and robustness. 
The image Jacobian can be estimated from an offline dataset which includes robot trajectories $D_x =\{ \mathbf{(x,v)}^n_{0:T},n=1,...,N\}$ and corresponding image feature trajectories $D_s =\{ \mathbf{(s,\dot{s})}^n_{0:T},n=1,...,N\}$. 
The estimated image Jacobian aims to approximate $\mathbf{\dot{s}} \simeq \mathbf{\hat{J}}_s \mathbf{v}$.
The target of estimating the image Jacobian can be described as
\begin{equation}
\hat{J}_s(\mathbf{s}) \Big|_{\mathbf{v} = \mathbf{v}_m} = \arg \min_{\mathbf{\hat{J}}_s} \sum_{m: \mathbf{v}_m \in D_x} \rho(\mathbf{\dot{s}}_m - \mathbf{\hat{J}}_s(\mathbf{s}_m) \mathbf{v}_m) \, ,
\end{equation}
where the $\rho(e) = e^2$ is a least-squares estimator.

Previous research highlighted the importance of fusing local Jacobian estimation with the global position information~\cite{VS_KNN-LLS}. 
We used a Gaussian Mixture Model (GMM) to represent global relationships and a Local Least Squares (LLS) estimator to approximate the local Jacobian matrix.

GMM uses a self-supervised probabilistic model to cluster the data to determine which local region the data sample belongs to~\cite{GMM}. 
The optimisation of the GMM is achieved by the Expectation-Maximisation (EM) algorithm. The dataset was clustered into $K$ different local regions. A binary indicator $z_i = [z_{i,1},...,z_{i,K}]^T$ was assigned to each data point $\boldsymbol{x}_i$ to indicate which cluster it belongs to.
In each of the regions, an LLS estimator was used to fit the mapping from the image plane space to Cartesian space as:
\begin{equation}
  \hat{f}_k(\mathbf{\dot{s}}_i | \mathbf{X}_k) = \arg \min_{\hat{f}_k} \sum_{i: \mathbf{\dot{s}}_i} \prod_{k: z_{i,k}} \rho( \mathbf{v}_i - \hat{f}_k (\mathbf{\dot{s}}_i) ) ^{z_{i,k}} \, ,
\end{equation}
where $\mathbf{v}_i$ is the corresponding Cartesian velocity at $\mathbf{\dot{s}}_i$ and $\mathbf{X}_k$ is the fitted linear weight. This weight is the estimated inversed image Jacobian according to its definition $\mathbf{v} = {\hat{\mathbf{J}^+_s}}  \mathbf{\dot{s}}$.

Using the fitted GMM could calculate the probability that a new data point $\mathbf{s}$ belongs to any of the clusters denoted as $\{ p( z_{k}=1 | \mathbf{s}) | k=1,...,K \}$. To fuse this global region information with the local estimation, a softmax is performed on the probability to obtain the global weight for mixing each local estimation. The resulting mixed image Jacobian is
\begin{equation}
    \hat{\mathbf{J}^+_s}(\mathbf{s}) = \sum_{k=1}^K \frac{\exp\left(p(z_{k}=1 \mid \mathbf{s})\right)}{\sum_{j=0}^K \exp\left(p(z_{j}=1 \mid \mathbf{s})\right)} \mathbf{X}_k \, ,
\end{equation}
which allows a globally smooth $\hat{\mathbf{J}^+_s}(\mathbf{s})$.

To test its validity, the estimated image Jacobian from this GMM-LSS method was compared with the analytical method~\cite{VS_tutorial} and a KMeans-LSS similar to the proposed method. 
The analytical method always failed, while the modified KMeans-LLS method failed occasionally as it frequently became stuck and oscillated at the decision boundary of clusters. 
The proposed GMM-LSS can successfully reach the target point and continuously transition to each cluster.

\subsection{Contact Height Estimation}
Precise control of the DRS probe's contact with the specimen is crucial for acquiring high-quality spectra. Visual servoing was initially used to manage the contact point, but it failed after the probe made contact with the specimen. This was caused by the intense light around the probe, which led to camera overexposure. This also made the stereo camera ineffective for depth estimation in this situation.

\begin{figure}[t]
\includegraphics[width=\linewidth]{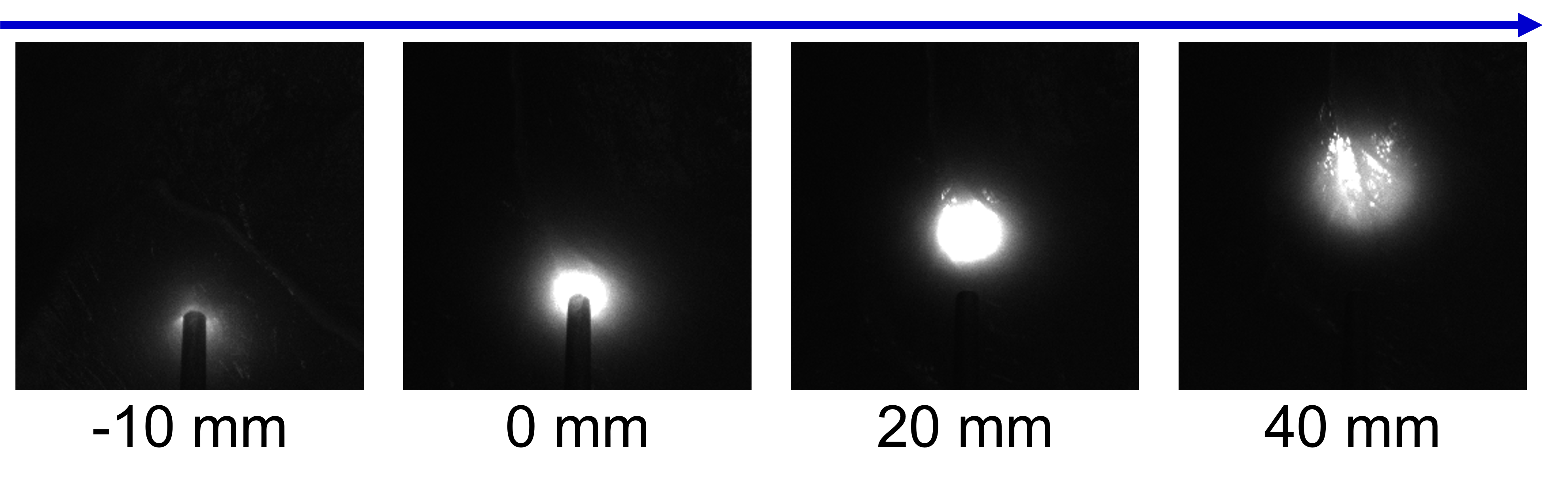}
\vspace{-25pt}
\caption{Example of wrist grey-scale camera images at different contact height. The width of the black background view in the camera view was $67\,\text{mm}$ when making contact at $0\,\text{mm}$.}
\label{fig:depth_figure_examples}
\end{figure}

The force sensor is unreliable due to tissue deformation, including lateral forces caused by horizontal displacement as the probe presses and advances.
Therefore, an image-based contact estimation algorithm was developed based on the shape of the illuminated light captured from the wrist camera. 
Its aperture is small enough to distinguish the shape change of reflected light according to the relative height as shown in \autoref{fig:depth_figure_examples}.
A neural network was trained to predict the current height, which offered superior performance and robustness compared to rule-based computer vision techniques.

A dataset was collected to train the neural network.
Before data collection, the robot was configured to always point straight down orthogonal to the sampling platform and was manually reset to the optimal contact plane (as $0\,\text{mm}$ in \autoref{fig:depth_figure_examples}) under the robot hand-guided mode. During data collection, the robot was controlled to move vertically upward for $40\,\text{mm}$ and down for $10\,\text{mm}$ from the optimal contact plane. The relative displacements, the wrist camera images, and the DRS measurements were recorded as the ground truth. A rump steak was used as a sample for data collection. A total of 45 episodes including 23 upward movements and 22 downward movements were collected. 
The network consisted of a ResNet-50 encoder and a linear head to predict the height, which was trained with Huber loss. 

\begin{figure}[t]
\includegraphics[width=\linewidth]{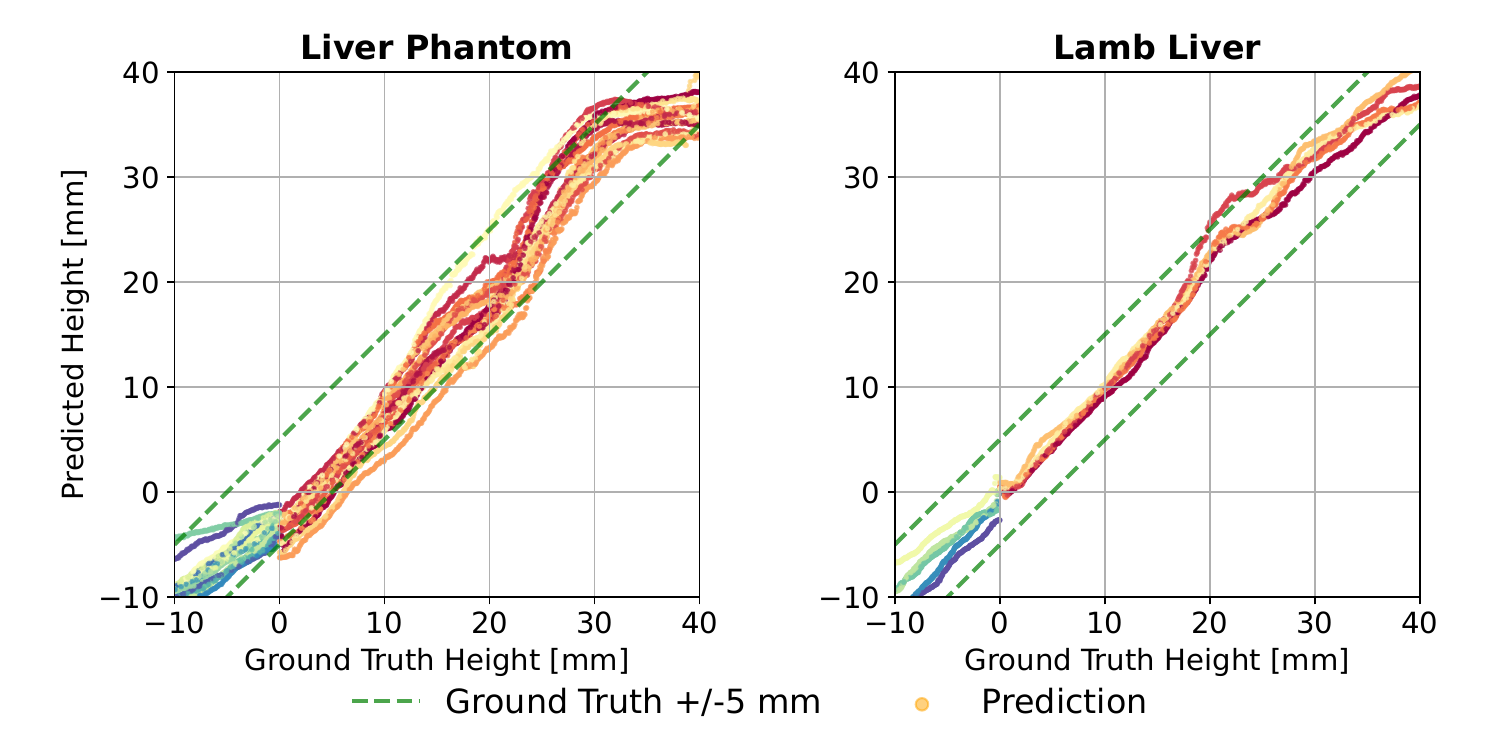}
\vspace{-20pt}
\caption{Visualising the evaluation results of height estimator neural network on testing dataset of horizontal movement on live phantom (10 trials) and lamb liver (30 trials). The range of $\pm5\,\text{mm}$ from the ground truth value is marked by green lines. Different trials are marked with different colors.}
\label{fig:height_estimator_eval}
\end{figure}

The testing results on unseen samples can be seen in \autoref{fig:height_estimator_eval}. It is shown that the system could successfully estimate the height of the lamb liver with a minor variance. For the liver phantom, a higher variance and error were observed around $0\,\text{mm}$, likely due to the heterogeneity of its surface scattering properties. However, this variance is generally acceptable, as the overall trend is well-represented, and the error can be corrected with an additional bias.

Estimating the contact height via the DRS spectrum might not be feasible. 
A further experiment utilised a 1D ResNet, which was trained to estimate the contact height based on known input spectra, showing a significantly higher variance with a weaker trend and generalisation difficulties in prediction compared to the image-based estimator. The Mean Absolute Error of this model is $1.12\,\text{mm}$, significantly higher than $0.06\,\text{mm}$ with the proposed image-based estimator.

\subsection{Motion Mixture}
A motion mixture module was built to blend the action from visual servoing $\mathbf{a}_{VS}$ and the contact height compensator $\mathbf{a}_{HC}$ for a balanced motion. 
This module included two PID controllers and a blending function as shown in \autoref{fig:main_fig}. 
The blending function assigns weights to $\mathbf{a}_{VS}$ and $\mathbf{a}_{HC}$, which are defined as: $\beta = \alpha + (1-\alpha)\times(1-e^{-\frac{d}{k}})\,$ where $\alpha$ is the lower limit of the weight for visual servoing, $d=h(t)-h^*$ is the distance error to the settle height $h^*$, and $k$ is a weight to adjust the response to distance error. 
Therefore, the blended action became $\mathbf{a} = \beta \mathbf{a}_{VS} + (1-\beta) \mathbf{a}_{HC}$.

\section{EXPERIMENTS}

\subsection{DRS Equipment and Scanning Task Setup}

The DRS was set up before each use. 
The spectrometer was calibrated using a standard reflector, Spectralon, to acquire the white $\mathbf{\mu}_{W}$ and dark field $\mathbf{\mu}_{D}$ reading with the light source powered on and off correspondingly. The reading was thus normalised as $\mathbf{\mu}_{DRS}=\frac{\mathbf{\mu}_{R}-\mathbf{\mu}_{D}}{\mathbf{\mu}_{W}-\mathbf{\mu}_{D}}$ given the raw signal reading $\mathbf{\mu}_{R}$. $\mathbf{\mu}_{DRS}$ was then processed using a Savitzky–Golay filter to perform noise reduction and preserve higher-order moments representing spectral characteristics. The visible light wavelength range from $468\,\text{nm}$ to $720\,\text{nm}$ of $\mathbf{\mu}_{DRS}$ was selected to represent the spectrum of the sample according to clinical practice~\cite{Nazarian2022JAMA}.

\subsection{System Evaluation}
In this experiment, the system was commanded to perform line scanning on four types of samples, including the stomach and liver sections of a silicone abdominal organ phantom, rump steak, and lamb liver. Each section of the phantom is deformable and made of homogeneous material on the surface but consists of different layers underneath. 
All meat samples were purchased from a local grocery store.

Before robotic sampling, a human operator first manually sampled the spectrum over the surface of the target object. The operator was instructed to scan the surface with minimal force while maintaining firm contact.
Each manual scan covered the selected region in approximately $30\,s$ and was repeated five times for each target object.

During the robotic sampling, a scanning trajectory was planned based on a straight line indicated by the user on the screen. 
The sampling procedure was repeated 12, 10, 15, and 25 times on four target objects correspondingly. 
The parameters of the controller were tuned to ensure an optimal contact condition. 
Empirically tuning of controller parameters suggested that adjusting the settled contact height $h^*$ and visual servoing motion weight $\alpha$ could efficiently reach the optimal contact condition. They were set to $0\,\text{mm}, -2\,\text{mm}, 0\,\text{mm}, 1\,\text{mm}$ to calibrate the height estimator and $0.2, 0.3, 0.4, 0.2$ to prevent aggressive motion on deformable tissue.

\section{RESULTS}
\subsection{Initial Approach Trajectory Evaluation}
\begin{figure}[t]
\includegraphics[width=\linewidth]{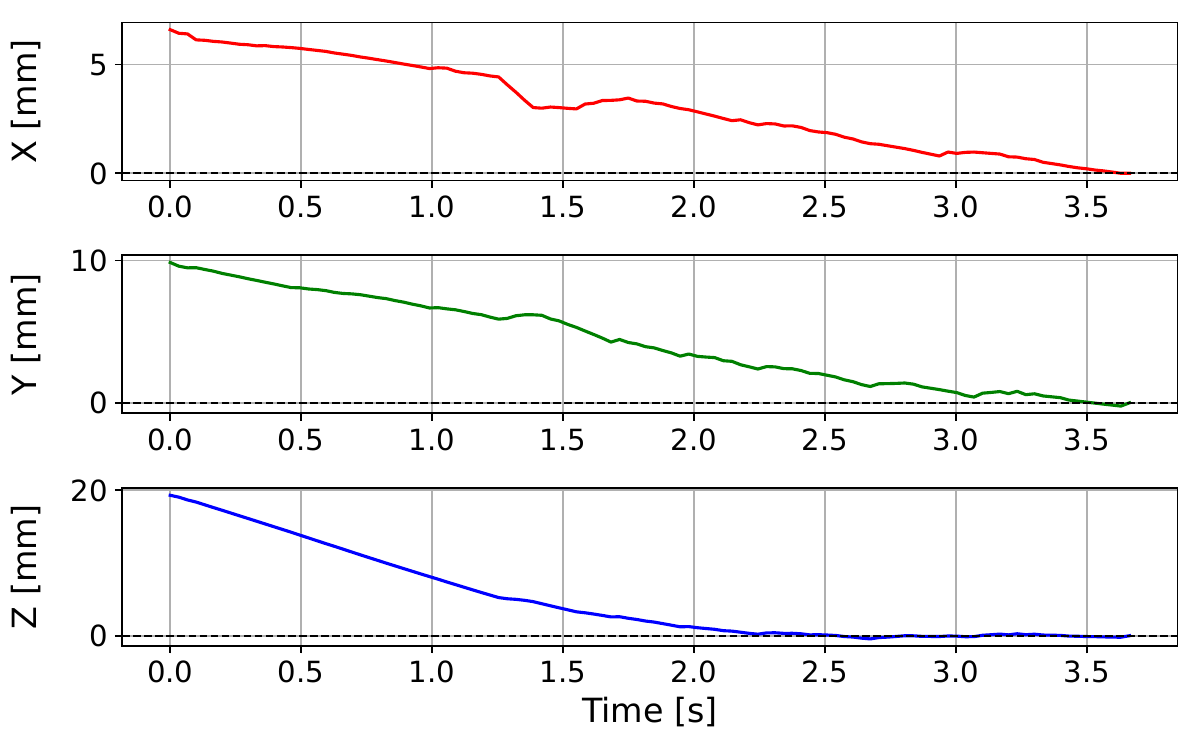}
\vspace{-25pt}
\caption{3D trajectory of approach to initial sampling point on a lamb liver sample. Positions are plotted by fixing target position at zero marked with a horizontal line.} 
\label{fig:trajectory_approaching}
\end{figure}

An example trajectory of the approach stage can be seen in \autoref{fig:trajectory_approaching}. It shows that the robot can effectively move from the initial point to the start point of the following scanning path. The robot tended to first move the probe to the contact plane with significant height change (Z-dimension) and then to the initial point skimming across the surface (X-Y dimension). This behaviour can be tuned by adjusting $\alpha$ and $k$ in the motion mixture. 
Although the task time is related to its distance from the initial point, the median time spent on all approach stages was $3.73\,\text{s}$ while $90\%$ of those were finished within $5.13\,\text{s}$. According to the preset scanning initialisation criteria, the starting point could be less than $1.29$ pixels (px) from the target point shown on the camera view. Manual measurement of the approach stage carried on the phantom resulted in position error $\leq2\,\text{mm}$ and contact height error $\leq1\,\text{mm}$. 
To validate the effectiveness of the height compensation module, it was disabled and the system was run 10 times. None of these trials was capable of reaching the standard contact height but either remained more than $5\,\text{mm}$ above the surface, producing unusable spectra, or high contact pressure, producing over-compressed contact which could damage tissue.  
These results show that the developed system could effectively initialise the scanning procedure by positioning the probe to the starting point with proper contact conditions.


\begin{table}[b]
\centering

\caption{Scanning trajectory image plane precision and Cartesian speed on each sample}

\resizebox{\columnwidth}{!}{

\begin{threeparttable} 

\begin{tabular}{lccccccc}
\toprule
\multirow{3}{*}{\textbf{Sample}} & \multicolumn{4}{c}{\textbf{Err. Distance [$px$]}} & \multicolumn{2}{c}{\textbf{C. Speed [$mm/s$]}} \\ \cmidrule(lr){2-5} \cmidrule(lr){6-7}
 & \multicolumn{2}{c}{\textbf{AVG}$\downarrow$} & \multicolumn{2}{c}{\textbf{.90}$\downarrow$} & \multirow{2}{*}{\textbf{AVG}$\uparrow$} & \multirow{2}{*}{\textbf{STD}$\downarrow$} \\ 
 \cmidrule(lr){2-3} \cmidrule(lr){4-5} 
 & \textbf{P} & \textbf{L} & \textbf{P} & \textbf{L} 
 \\
 \midrule
Liver Phantom   & 4.97 & 3.21 & 9.75 & 6.65 & 6.04 & 0.84 \\ 
Stomach Phantom & 2.62 & 3.60 & 4.68 & 7.12 & 5.00 & 0.69 \\ 
Rump Steak      & 2.96 & 3.31 & 5.96 & 6.25 & 4.66 & 0.69 \\ 
Lamb Liver      & 3.56 & 3.74 & 6.88 & 6.42 & 4.41 & 0.75 \\ \bottomrule
\end{tabular}

\begin{tablenotes}[flushleft]
\footnotesize
\item The (\textbf{Err. Distance}) shows the mean (\textbf{AVG}) and the $90^{th}$ Percentile (\textbf{.90}) of the pixel error distance from current sampling point, represented by probe tip (\textbf{P}) and light centre (\textbf{L}) to the desired line scanning trajectory on image plane. The \textbf{C. Speed} shows the mean (\textbf{AVG}) and standard deviation (\textbf{STD}) of the scanning speed in Cartesian space.
\end{tablenotes}

\end{threeparttable}
}

\label{tab:trajectory_metric}

\end{table}
\begin{figure}[tb]
\includegraphics[width=\linewidth]{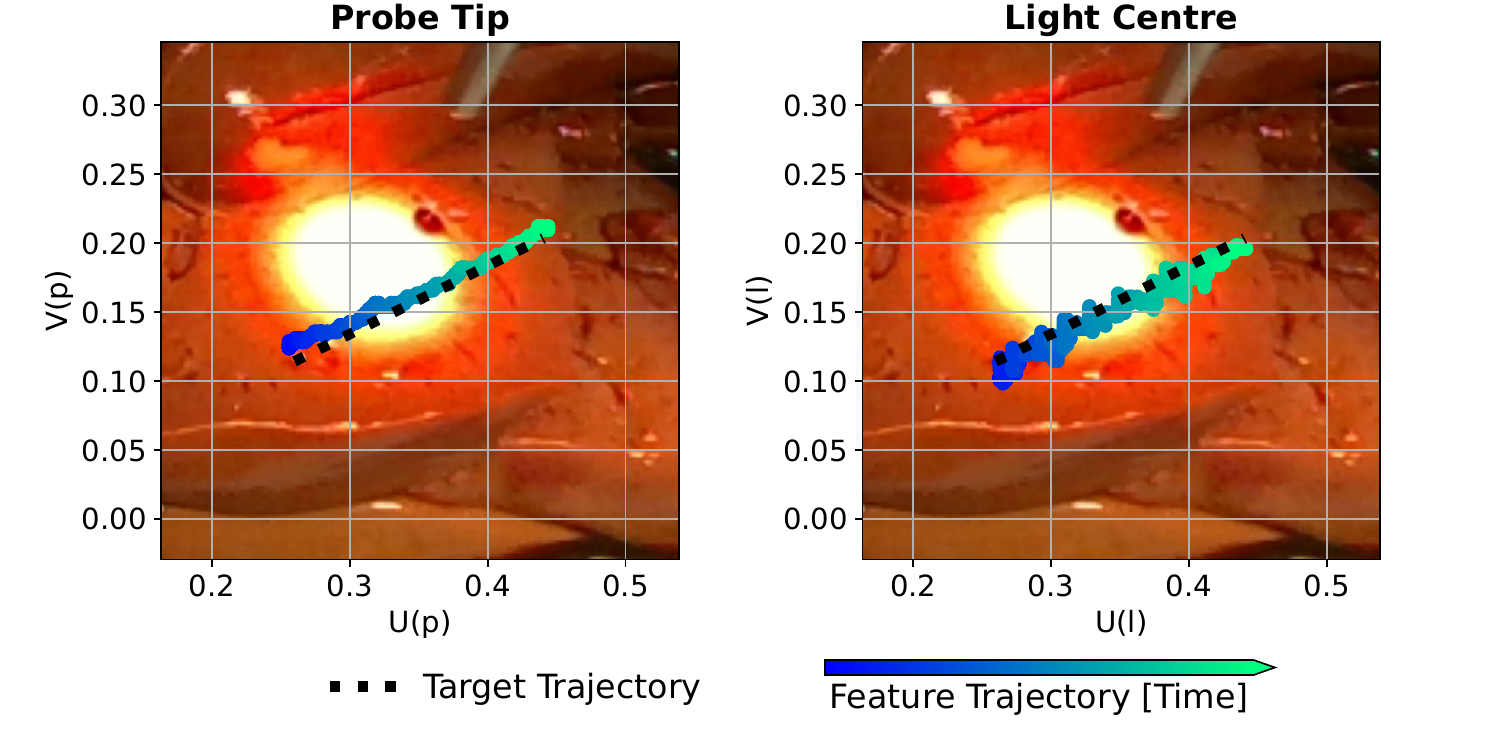}
\vspace{-25pt}
\caption{Visualisation of image plane trajectory of two image features overlaid on the zoomed-in third-person-view camera image (captured at the beginning) during in-contact automatic scanning of a lamb liver sample.} 
\label{fig:trajectory_uv_plane}
\end{figure}
\subsection{Scanning Trajectory Evaluation}

\begin{figure}[t]
\includegraphics[width=\linewidth]{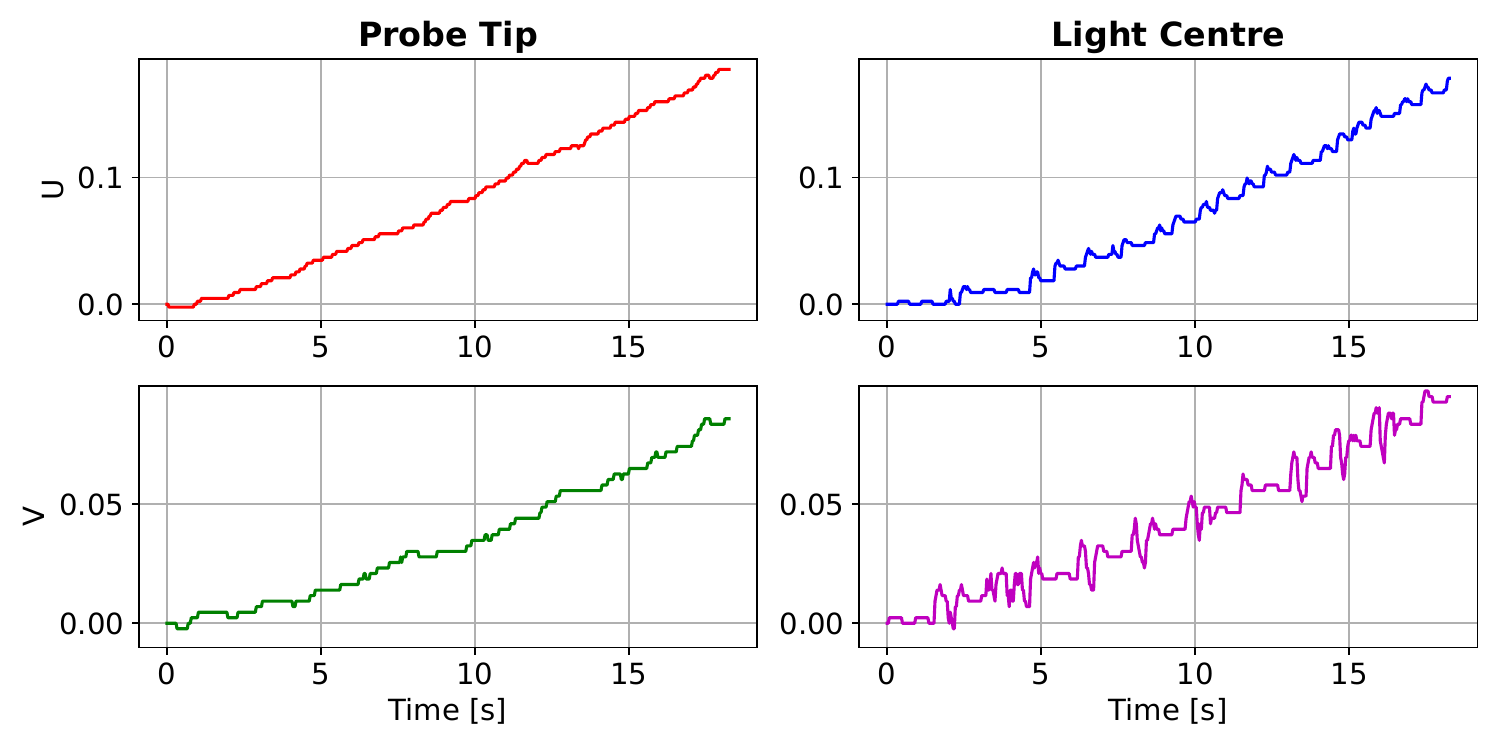}
\vspace{-20pt}
\caption{Image plane trajectory of image features, probe tip ($u_t$ and $v_t$) and illuminated light centre ($u_l$ and $v_l$), during in-contact automatic scanning of a lamb liver sample.} 
\label{fig:trajectory_scanning}
\end{figure}
\begin{figure*}[htb]
    \centering
    \includegraphics[width=\linewidth]{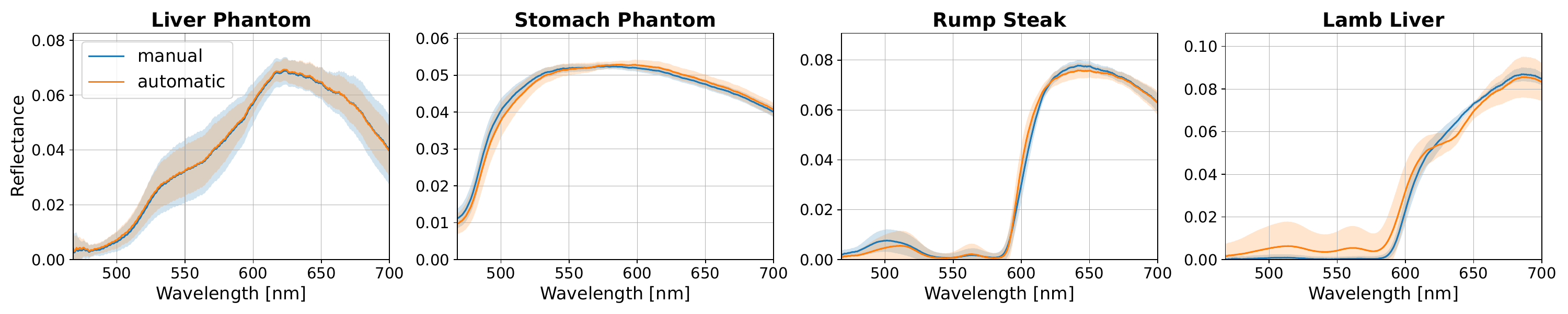}
    \includegraphics[width=\linewidth]{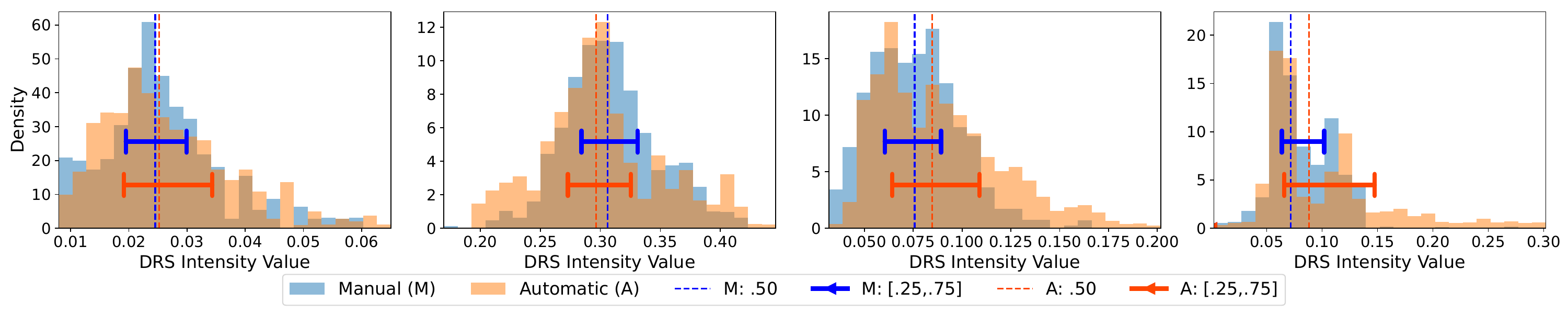}
    \vspace{-20pt}
    \caption{\textbf{Visualisation of collected DRS fingerprint and intensity.} Metrics for DRS spectrum collected from hand-held manual collection and robotic-assisted automatic collection on four type of objects: a) mean of DRS spectrum fingerprint with standard deviation marking its range, b) histogram of DRS intensity distribution for manual (M) and automatic (A) collected spectrum with $25^{th}$, $50^{th}$ (Median), $75^{th}$ percentile location marked}
    \label{fig:fingerprint_and_intensity}
\end{figure*}
The in-contact scanning can be successfully achieved in more than $85\%$ of trials. The failure cases were mainly due to unsuccessful feature detection of the light centre position. A visualisation of the trajectory is shown in \autoref{fig:trajectory_uv_plane}. It shows that the robot could follow the line scanning command indicated on the camera view. The scanning trajectory of two image features for a successful scanning is shown in \autoref{fig:trajectory_scanning}. However, an oscillation could be observed for the light centre trajectory compared to the corresponding probe tip trajectory. This could be because of the difference in sensitivity between the two control branches. The scanning trajectory tracking error on the image plane is shown in \autoref{tab:trajectory_metric}. The average error distance measured in pixel could be less than $5\,\text{px}$ with $90\%$ error less than $10\,\text{px}$. The scanning speeds were consistent in different cases and generally matched the manual scanning speed for clinical sampling.

\begin{table}[b]
\centering

\caption{consistency of manual and automatic collection spectrum}

\begin{threeparttable} 

\begin{tabularx}{\linewidth}{lXXXXX}
\toprule
\textbf{Sample} & \textbf{RMSE}$\downarrow$ & $\mathbf{\theta}\downarrow$ & $\sigma_{M}\downarrow$ & $\sigma_{A}\downarrow$ \\ \midrule
Liver Phantom    & 0.542   & 0.229       & 2.36        & \textbf{2.29}       \\ 
Stomach Phantom  & 1.288    & 0.073        & \textbf{0.98}        & 1.07       \\ 
Rump Steak       & 2.108      & 0.119        & \textbf{3.32}         & 3.35      \\ 
Lamb Liver       & 4.406     & 0.222        & 3.51        & \textbf{3.32}      \\ \bottomrule

\end{tabularx}

\begin{tablenotes}[flushleft]   
\footnotesize
\item \textbf{RMSE} ($10^{-3}$ as the base) is the Root Mean Square Error of automatic collected $\mathbf{\mu}_f[A]$ comparing to manual collected $\mathbf{\mu}_f[M]$. $\mathbf{\theta}$ is the spectral angle between $\mathbf{\mu}_f[A]$ and $\mathbf{\mu}_f[M]$. $\sigma_{A}$ and $\sigma_{M}$  (both $10^{-2}$ as the base) are the standard deviation of corresponding spectral fingerprints.
\end{tablenotes}

\end{threeparttable}

\label{tab:fringerprint_and_intensity_metric}

\end{table}

\subsection{Spectrum Evaluation}

The spectra collected from scanning on four different types of tissues were evaluated and compared regarding their spectral fingerprint $\mathbf{\mu}_f$ and intensity distribution $\mathbf{\mu}_i$ with manually collected spectra. The visualisation of $\mathbf{\mu}_f$ and $\mathbf{\mu}_i$ is shown in \autoref{fig:fingerprint_and_intensity}. For $\mathbf{\mu}_f$, the shape is generally identical with a similar variance range comparing the automatic collection with the manual collection. The mismatching of $\mathbf{\mu}_f$ shape is more significant in the meat sample compared to the silicone sample. For $\mathbf{\mu}_i$, the distribution is similar with larger overlap but $\mathbf{\mu}_i[M]$ from manual collection is more concentrated with lower intensity than $\mathbf{\mu}_i[A]$ from automatic collection. The percentile threshold is used to describe the width of such a skewed distribution. 

The quantitative comparison of $\mathbf{\mu}_f[M]$ and $\mathbf{\mu}_f[A]$ is shown in \autoref{tab:fringerprint_and_intensity_metric}. 
The difference between them depends on the properties of the target object, possibly due to the heterogeneity of the target sample, and is more pronounced in animal meat with anatomical structures than in the human-made phantom.
From the standard deviations of spectral fingerprint, the automatic collection could lead to lower standard deviation in two types of target object, indicating a compatible intra-observer variance than manual collection.

\section{DISCUSSION AND CONCLUSION}\label{discussion}
In this work, we propose an automatic robotic-assisted DRS scanning system. It utilises visual servoing with an estimated image Jacobian to execute scanning commands. 
An image-based height compensation module was developed to precisely detect the contact height and consistently maintain an optimal contact condition. 

Trajectory analysis indicates that it could follow the scanning command within $5\,\text{px}$ error. 
Spectrum analysis shows that automatic robot scanning leads to comparable results in terms of spectral fingerprints with small variance compared to manual scanning.
This work is the first to achieve automatic DRS scanning with contact management, eliminating the need for invasive tissue preparation. This represents a step toward integrating intra-operative tissue assessment with DRS into clinical practice.

The current system lacks control over probe orientation, which may lead to small gaps when making contact. 
Such conditions can introduce specular reflections, causing alterations in the measured spectrum. Therefore, incorporating orientation control could further enhance the consistency of the acquired spectral data. Additionally, further experiments on a broader range of tissue specimens are necessary to validate the precision and generalisability of this system. 


\section*{ACKNOWLEDGMENTS}
This paper is independent research funded by the National Institute for Health Research (NIHR) Imperial Biomedical Research Centre (BRC), the Cancer Research UK (CRUK) Imperial Centre, the Wellcome Trust ITPA MedTechOne awards. 
The authors would like to thank Yuxuan Shu, Tobias Czempiel, Jacopo Hu, Junhong Chen, Ioannis Gkouzionis, Maxime Giot, and Edward Johns for their insightful discussions, and Fernando Bello for providing the silicone abdominal phantom.

\newpage
\bibliographystyle{IEEEtran}
\bibliography{IEEEabrv,icra_25_reference}

\end{document}